\documentclass[10pt,twocolumn,letterpaper]{article}
\usepackage{array,multirow,graphicx}
\usepackage{cvpr}
\usepackage{times}
\usepackage{epsfig}
\usepackage{graphicx}
\usepackage{amsmath}
\usepackage{amssymb}
\usepackage{stmaryrd}
\usepackage{dsfont}

\usepackage[breaklinks=true,bookmarks=false]{hyperref}

\cvprfinalcopy 

\def\cvprPaperID{****} 

\newcommand{\bs}[1]{\ensuremath{\boldsymbol{#1}}}

\begin{document}

\title{A Simple Loss Function for Improving the Convergence and Accuracy of Visual Question Answering Models}

\author{
Ilija Ilievski\\
  Integrative Sciences and Engineering \\
  National University of Singapore \\
{\tt\small ilija.ilievski@u.nus.edu}
\and
Jiashi Feng\\
Electrical and Computer Engineering\\
National University of Singapore\\
{\tt\small elefjia@nus.edu.sg}
}

\maketitle

\begin{abstract}
   Visual question answering as recently proposed multimodal learning task has enjoyed wide attention from the deep learning community.
   Lately, the focus was on developing new representation fusion methods and attention mechanisms to achieve superior performance. 
   On the other hand, very little focus has been put on the models' loss function, arguably one of the most important aspects of training deep learning models. 
   The prevailing practice is to use cross entropy loss function that penalizes the probability given to all the answers in the vocabulary except the single most common answer for the particular question.
   However, the VQA evaluation function compares the predicted answer with all the ground-truth answers for the given question and if there is a matching, a partial point is given.
   This causes a discrepancy between the model's cross entropy loss and the model's accuracy as calculated by the VQA evaluation function.
   In this work, we propose a novel loss, termed as soft cross entropy, that considers all ground-truth answers and thus reduces the loss -- accuracy discrepancy.
   The proposed loss leads to an improved training convergence of VQA models and an increase in accuracy as much as $1.6\%$.
\end{abstract}
\vspace{-1em}
\section{Introduction}

Visual question answering (VQA) requires an AI agent to answer questions about an image. 
As a challenging multimodal problem and a proxy task for visual reasoning, it has attracted a lot of attention from the deep learning community. 
Multiple models were introduced~\cite{hu2017learning,santoro2017simple,johnson2017inferring} and a new dataset with a specific focus on visual reasoning~\cite{johnson2016clevr}.

The currently largest VQA dataset, VQA v2.0~\cite{balanced_vqa_v2}, contains $1.1$ million questions for the $205$ thousand MS COCO images~\cite{lin2014microsoft}.
Each question is paired with ten human-provided answers.
The usual VQA model uses a pretrained ResNet~\cite{he-15} network to obtain an image representation and an LSTM~\cite{hochreiter-97} unit to learn a representation of the question words.
The model then fuses the two representations into a single multimodal representation via element-wise multiplication or other more sophisticated methods. 
Finally, the most common answer out of the ten provided is used to train the model to classify the multimodal representation to a correct answer~\cite{yang-15,ilievski2016focused,xu-15,shih-15,xiong2016dynamic}.

Recently, several representation fusion methods were developed~\cite{fukui2016multimodal,kim2016hadamard,benyounescadene2017mutan} and some novel attention mechanisms were introduced~\cite{nam2016dual,lu2016hierarchical}.
But, very little attention has been put on the VQA model loss function, which is an essential part of its training. 
   The prevailing approach is to use the most common answer and a cross entropy loss function (Eq.~\eqref{NLL}).
   However, a VQA model is evaluated by comparing the predicted answer with {\em all} the ground-truth answers for a given question and if there is a match, a partial point is given.
   This causes a discrepancy between the model's cross entropy loss and the model's accuracy as calculated by the VQA evaluation function, which in turn results in a delayed training convergence and reduced test accuracy.

   In this work, we propose a new loss function, termed as {\em soft} cross entropy, that considers {\em all} ground-truth answers and thus solves the discrepancy problem. 
   In contrast to the standard cross entropy loss, the soft cross entropy loss provides to the model a set of plausible answers for a given question and information about the question's ambiguity.
   As a consequence, the VQA models trained with the proposed loss have a stable training process, converge faster, and achieve on average $1.5\%$ higher accuracy than models trained with the standard cross entropy loss function.

   In summary, the contributions of this work are:
   \begin{itemize}
         \item We propose a novel loss function for VQA, that more closely reflects a VQA model's performance. The proposed loss is justified with error analysis and empirical evaluation. 
         \item We provide an efficient code for reproducing the experimental results and to serve as a starter code to the VQA community.
   \end{itemize}

\begin{table}[t]
   \centering
   \caption{Best validation accuracy  on the VQA v2.0 validation set.}
   \begin{tabular}{|p{.05cm}|l|cccc|}
      \hline
       \parbox[t]{2mm}{{{}}} & 
       Loss Function                    &All&Y/N&Num&Other\\
      \hline
      \parbox[t]{0mm}{\multirow{2}{*}{\rotatebox[origin=c]{90}{\footnotesize AVG}}} & 
      Cross Entropy               &46.8&55.8&29.8&42.4\\
      &      Soft Cross Entropy           &\bs{48.0}&\bs{57.1}&\bs{31.0}&\bs{43.3}\\
      \hline
      \parbox[t]{1mm}{\multirow{2}{*}{\rotatebox[origin=c]{90}{\footnotesize POOL}}}&
      Cross Entropy              &58.8&70.1&37.5&53.1\\
      &    Soft Cross Entropy          &\bs{60.4}&\bs{71.9}&\bs{39.0}&\bs{54.6}\\
      \hline
   \end{tabular}
   \label{tbl:vqa2}
\vspace{-1.2em}
\end{table}
\vspace{-1em}
\begin{figure}[ht]
   \includegraphics[height=.6\linewidth,width=\linewidth]{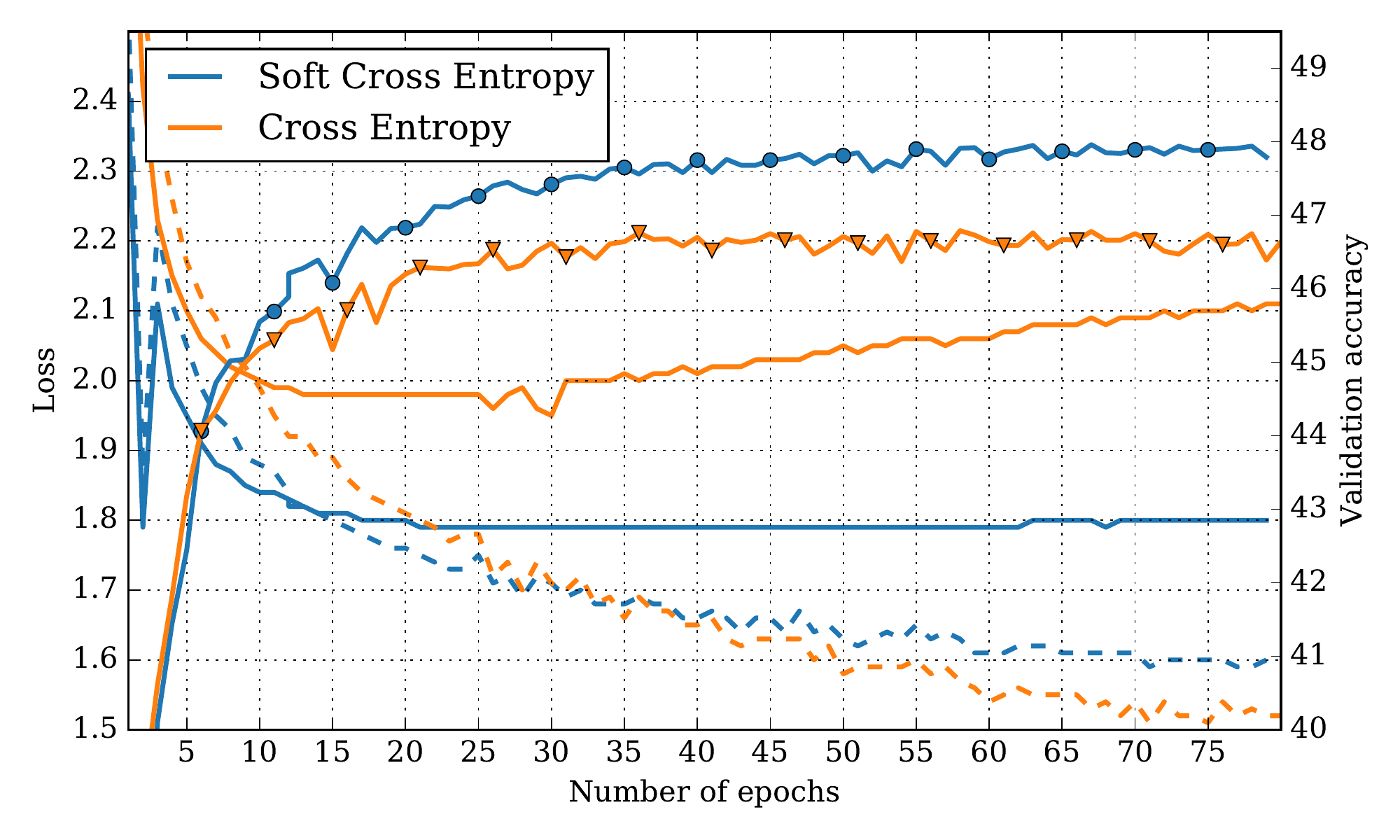}
   \caption{Training (dashed lines) and validation (solid lines) loss and validation accuracy (dotted lines) for the \textbf{AVG} VQA model.}
   \label{fig:cross1}
\vspace{-1em}
\end{figure}
\begin{figure}[ht]
   \includegraphics[height=.6\linewidth,width=\linewidth]{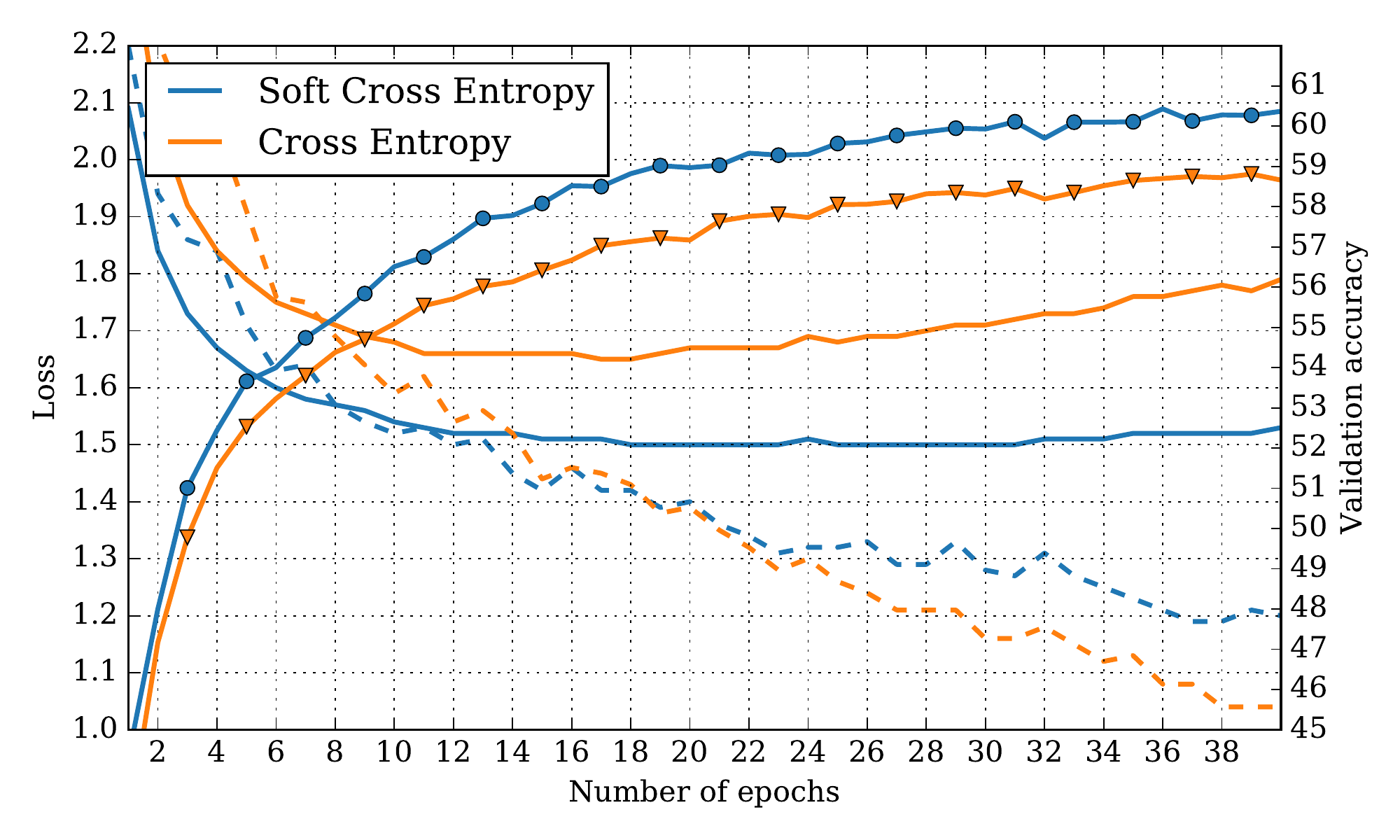}
   \caption{Training (dashed lines) and validation (solid lines) loss and validation accuracy (dotted lines) for the \textbf{POOL} VQA model.}
\vspace{-1.0em}
   \label{fig:cross2}
\end{figure}
\vspace{-0.0em}
\section{Soft Cross Entropy Loss}
\vspace{-0.3em}
The VQA problem can be reduced to a maximum likelihood estimation problem, where the model classifies a question-image pair to an answer from the training set.
Generally, a deep learning model is trained on a classification problem using a cross entropy loss function:
\vspace{-0.7em}
\begin{equation}
   \mathcal{L}({\bs{x}}, c^*) = -\bs{x}_{c^*} + \log\big(\sum_{j=1}^{|\bs{x}|} \exp(\bs{x}_{j})\big),
   \label{NLL}
\vspace{-0.7em}
\end{equation}
where \bs{x} is a vector of network activations for each class and $c^*$ is the index of the correct class.

However, contrary to conventional classification problems, the VQA evaluation metric considers a predicted answer as correct if the answer was given by at least three out of ten human annotators.
The accuracy is then averaged over all ${10 \choose 9}$ subsets of ground-truth answers:
\begin{equation*}
   Acc(a) = \frac{1}{10} \sum_{k = 1}^{10} \min(\frac{\sum_{j = 1, j \neq k}^{10}\mathds{1}(a = a_j)}{3}, 1).
\vspace{-0.5em}
\end{equation*}

As a result, the model's performance is not properly assessed with the cross entropy loss function during the training phase.
The improper loss function has significant negative impact on the model's training and delays the convergence.
Furthermore, it results in abnormal and counter-intuitive validation loss -- accuracy relationship where both the loss and the accuracy increase (Fig.~\ref{fig:cross1} and~\ref{fig:cross2}).

As a solution, we propose to use a loss function that considers {\em all} ground-truth answers.
The proposed loss function, termed as {\em soft} cross entropy, is a simple weighted average of each unique ground-truth answer:
\begin{equation}
   \begin{gathered}
        \mathcal{L}({\bs{x}},\bs{c}, \bs{w}) = \sum_{i=1}^{|\bs{c}|} w_i \Big(-\bs{x}_{c_i} + \log\big(\sum_{j=1}^{|\bs{x}|} \exp(\bs{x}_{j})\big)\Big), \\
   \end{gathered}
   \label{softNLL}
\end{equation}
where $\bs{c}$ is a vector of unique ground-truth answers and \bs{w} is a vector of answer weights computed as the number of times the unique answer appears in the ground-truth set divided by the total number of answers.

\vspace{-0.5em}
\section{Experiments}
\vspace{-0.2em}
We evaluate the proposed loss function on the recently released VQA v2.0 benchmark dataset~\cite{balanced_vqa_v2}.
To demonstrate the general applicability of the proposed loss we train a variant of the two most common VQA models~\cite{lu-15,kim2016hadamard}.

Both models use an LSTM to encode the question words to a vector representation. 
The \textbf{AVG} model is based on~\cite{lu-15} and it utilizes the activations of the penultimate layer of pretrained ResNet-152\cite{he-15} as image representation and does not employ attention mechanism.
The \textbf{POOL} model is based on~\cite{kim2016hadamard} and it considers the tensor of activations of the last pooling layer of the same ResNet and employs attention mechanism over the regions to obtain an image representation in a vector form.
Both models are trained with Adam~\cite{kingma2014adam} and a batch size of $64$\footnote{Code available at~\url{github.com/ilija139/vqa-soft}}.

\vspace{-0.5em}
\section{Discussion and Conclusion}
From Table~\ref{tbl:vqa2} we can observe that the proposed loss increases the overall accuracy by $1.2\%$ in the simpler model and $1.6\%$ increase in the pooling model.
The accuracy is increased for both models and all answer types which proves the general applicability of the soft cross entropy loss.

In Figures~\ref{fig:cross1} and \ref{fig:cross2} we can clearly observe the abnormal relationship between the validation loss and accuracy where they both start to increase near the half of the training process. 
Furthermore, we can observe how the cross entropy loss rapidly reduces on the training set without an increase in validation accuracy and a decrease in validation loss. 

The evaluation results show that by modeling the VQA evaluation metric more faithfully than conventional classification loss functions, the proposed loss function is able to bring a consistent increase in accuracy for VQA models.

{\small
\bibliographystyle{ieee}
\bibliography{vqa}

\begin{thebibliography}{10}\itemsep=-1pt

\bibitem{benyounescadene2017mutan}
H.~Ben-Younes, R.~Cad{\`{e}}ne, N.~Thome, and M.~Cord.
\newblock {MUTAN}: Multimodal {Tucker} fusion for visual question answering.

\bibitem{fukui2016multimodal}
A.~Fukui, D.~H. Park, D.~Yang, A.~Rohrbach, T.~Darrell, and M.~Rohrbach.
\newblock Multimodal compact bilinear pooling for visual question answering and
  visual grounding.
\newblock {\em arXiv preprint arXiv:1606.01847}, 2016.

\bibitem{balanced_vqa_v2}
Y.~Goyal, T.~Khot, D.~Summers{-}Stay, D.~Batra, and D.~Parikh.
\newblock Making the {V} in {VQA} matter: Elevating the role of image
  understanding in {V}isual {Q}uestion {A}nswering.
\newblock In {\em Conference on Computer Vision and Pattern Recognition
  (CVPR)}, 2017.

\bibitem{he-15}
K.~He, X.~Zhang, S.~Ren, and J.~Sun.
\newblock Deep residual learning for image recognition.
\newblock {\em arXiv preprint arXiv:1512.03385}, 2015.

\bibitem{hochreiter-97}
S.~Hochreiter and J.~Schmidhuber.
\newblock Long short-term memory.
\newblock {\em Neural computation}, 9(8):1735--1780, 1997.

\bibitem{hu2017learning}
R.~Hu, J.~Andreas, M.~Rohrbach, T.~Darrell, and K.~Saenko.
\newblock Learning to reason: End-to-end module networks for visual question
  answering.
\newblock {\em arXiv preprint arXiv:1704.05526}, 2017.

\bibitem{ilievski2016focused}
I.~Ilievski, S.~Yan, and J.~Feng.
\newblock A focused dynamic attention model for visual question answering.
\newblock {\em arXiv preprint arXiv:1604.01485}, 2016.

\bibitem{lu-15}
D.~B. Jiasen~Lu, Xiao~Lin and D.~Parikh.
\newblock Deeper {LSTM} and normalized {CNN} visual question answering model.
\newblock \url{https://github.com/VT-vision-lab/VQA_LSTM_CNN}, 2015.

\bibitem{johnson2016clevr}
J.~Johnson, B.~Hariharan, L.~van~der Maaten, L.~Fei-Fei, C.~L. Zitnick, and
  R.~Girshick.
\newblock Clevr: A diagnostic dataset for compositional language and elementary
  visual reasoning.
\newblock {\em arXiv preprint arXiv:1612.06890}, 2016.

\bibitem{johnson2017inferring}
J.~Johnson, B.~Hariharan, L.~van~der Maaten, J.~Hoffman, L.~Fei-Fei, C.~L.
  Zitnick, and R.~Girshick.
\newblock Inferring and executing programs for visual reasoning.
\newblock {\em arXiv preprint arXiv:1705.03633}, 2017.

\bibitem{kim2016hadamard}
J.-H. Kim, K.-W. On, J.~Kim, J.-W. Ha, and B.-T. Zhang.
\newblock Hadamard product for low-rank bilinear pooling.
\newblock {\em arXiv preprint arXiv:1610.04325}, 2016.

\bibitem{kingma2014adam}
D.~Kingma and J.~Ba.
\newblock Adam: A method for stochastic optimization.
\newblock {\em arXiv preprint arXiv:1412.6980}, 2014.

\bibitem{lin2014microsoft}
T.-Y. Lin, M.~Maire, S.~Belongie, J.~Hays, P.~Perona, D.~Ramanan,
  P.~Doll{\'a}r, and C.~L. Zitnick.
\newblock Microsoft {COCO}: Common objects in context.
\newblock In {\em European Conference on Computer Vision}, pages 740--755.
  Springer, 2014.

\bibitem{lu2016hierarchical}
J.~Lu, J.~Yang, D.~Batra, and D.~Parikh.
\newblock Hierarchical question-image co-attention for visual question
  answering.
\newblock In {\em Advances In Neural Information Processing Systems}, pages
  289--297, 2016.

\bibitem{nam2016dual}
H.~Nam, J.-W. Ha, and J.~Kim.
\newblock Dual attention networks for multimodal reasoning and matching.
\newblock {\em arXiv preprint arXiv:1611.00471}, 2016.

\bibitem{santoro2017simple}
A.~Santoro, D.~Raposo, D.~G. Barrett, M.~Malinowski, R.~Pascanu, P.~Battaglia,
  and T.~Lillicrap.
\newblock A simple neural network module for relational reasoning.
\newblock {\em arXiv preprint arXiv:1706.01427}, 2017.

\bibitem{shih-15}
K.~J. Shih, S.~Singh, and D.~Hoiem.
\newblock Where to look: Focus regions for visual question answering.
\newblock {\em arXiv preprint arXiv:1511.07394}, 2015.

\bibitem{xiong2016dynamic}
C.~Xiong, S.~Merity, and R.~Socher.
\newblock Dynamic memory networks for visual and textual question answering.
\newblock {\em arXiv}, 1603, 2016.

\bibitem{xu-15}
H.~Xu and K.~Saenko.
\newblock Ask, attend and answer: Exploring question-guided spatial attention
  for visual question answering.
\newblock {\em arXiv preprint arXiv:1511.05234}, 2015.

\bibitem{yang-15}
Z.~Yang, X.~He, J.~Gao, L.~Deng, and A.~Smola.
\newblock Stacked attention networks for image question answering.
\newblock {\em arXiv preprint arXiv:1511.02274}, 2015.

\end{thebibliography}
}

\end{document}